\documentclass[10pt,twocolumn,letterpaper]{article}
\usepackage{iccv}
\usepackage{times}
\usepackage{epsfig}
\usepackage{graphicx}
\usepackage{amsmath}
\usepackage{amssymb}
\usepackage{url}
\usepackage[normalem]{ulem}
\usepackage[table,xcdraw]{xcolor}
\usepackage{multirow}
\usepackage{amsthm}
\usepackage{mathrsfs}
\usepackage{bm}
\usepackage{lipsum}
\usepackage{subfigure}
\usepackage[section]{placeins}
\usepackage{booktabs}
\usepackage{xcolor}
\usepackage{cite}

\def \model {\textit{MVSS-Net}}

\definecolor{darkgreen}{RGB}{50,150,50}

\newcommand{\specialcell}[2][c]{%
  \begin{tabular}[#1]{@{}c@{}}#2\end{tabular}}
  
\usepackage[pagebackref=true,breaklinks=true,letterpaper=true,colorlinks,bookmarks=false]{hyperref}

\iccvfinalcopy 
\newcommand{\add}[1]{\textcolor{black}{#1}}
\newcommand{\del}[1]{\textcolor{purple}{\sout{}}}
\def\iccvPaperID{3997} 

\ificcvfinal\fi
\begin{document}
\title{Image Manipulation Detection by Multi-View Multi-Scale Supervision}

\author{Xinru Chen\textsuperscript{1,2*}, Chengbo Dong\textsuperscript{1,2*}, Jiaqi Ji\textsuperscript{1,2}, Juan Cao\textsuperscript{3,4}, Xirong Li\textsuperscript{1,2$\dagger$}\\
\textsuperscript{1}MoE Key Lab of Data Engineering and Knowledge Engineering, Renmin University of China\\
\textsuperscript{2}AIMC Lab, School of Information, Renmin University of China\\
\textsuperscript{3}Institute of Computing Technology, Chinese Academy of Sciences\\
\textsuperscript{4}State Key Laboratory of Media Convergence Production Technology and Systems
}

\maketitle
\thispagestyle{empty}

\newcommand\blfootnote[1]{%
\begingroup
\renewcommand\thefootnote{}\footnote{#1}%
\addtocounter{footnote}{-1}%
\endgroup
}

\begin{abstract}

The key challenge of image manipulation detection is how to learn generalizable features that are \emph{sensitive} to manipulations in novel data, whilst \emph{specific} to prevent false alarms on authentic images. Current research emphasizes the sensitivity, with the specificity overlooked. In this paper we address both aspects by multi-view feature learning and multi-scale supervision. By exploiting noise distribution and boundary artifact surrounding tampered regions, the former aims to learn semantic-agnostic and thus more generalizable features. The latter allows us to learn from authentic images which are nontrivial to be taken into account by current semantic segmentation network based methods. Our thoughts are realized by a new network which we term \emph{MVSS-Net}. Extensive experiments on five benchmark sets justify the viability of MVSS-Net for both pixel-level and image-level manipulation detection.
\end{abstract}

\section{Introduction}\label{sec:introduction}
\blfootnote{*Xinru Chen and Chengbo Dong contribute equally to this work.}
\blfootnote{$\dagger$Corresponding author: Xirong Li (xirong@ruc.edu.cn)}





Digital images can now be manipulated with ease and often in a visually imperceptible manner \cite{Gafni_2020_CVPR}. \emph{Copy-move} (copy and move elements from one region to another region in a given image), \emph{splicing} (copy elements from one image and paste them on another image) and \emph{inpainting} (removal of unwanted elements) are three common types of image manipulation that could lead to misinterpretation of the visual content \cite{JLSTM,Mahfoudi2019DEFACTO,review2}. This paper targets at auto-detection of images subjected to these types of manipulation. We aim to not only discriminate manipulated images from the authentic, but also pinpoint tampered  regions at the pixel level.

\begin{figure} 
    \begin{center}
    \includegraphics[width=0.94\columnwidth]{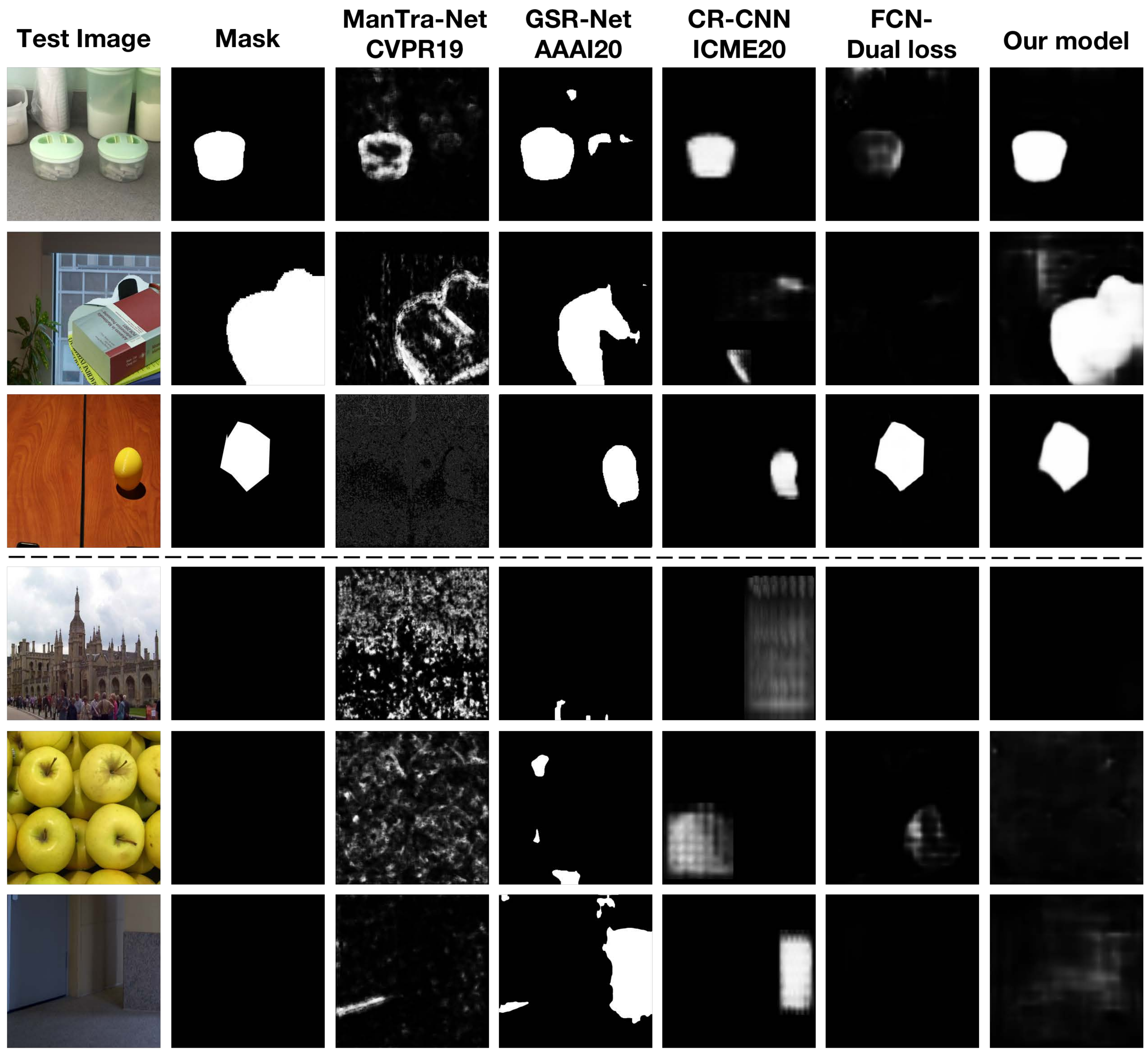}
    \end{center}
    \caption{\textbf{Image manipulation detection by the state-of-the-arts}. The first three rows are copy-move, splicing and inpainting, followed by three authentic images (thus with blank mask). Our model strikes a good balance between sensitivity and specificity.}
    \label{fig:intro-samples}
\end{figure}

\begin{figure*}[htpb]
    \begin{center}
    \includegraphics[width=0.93\textwidth]{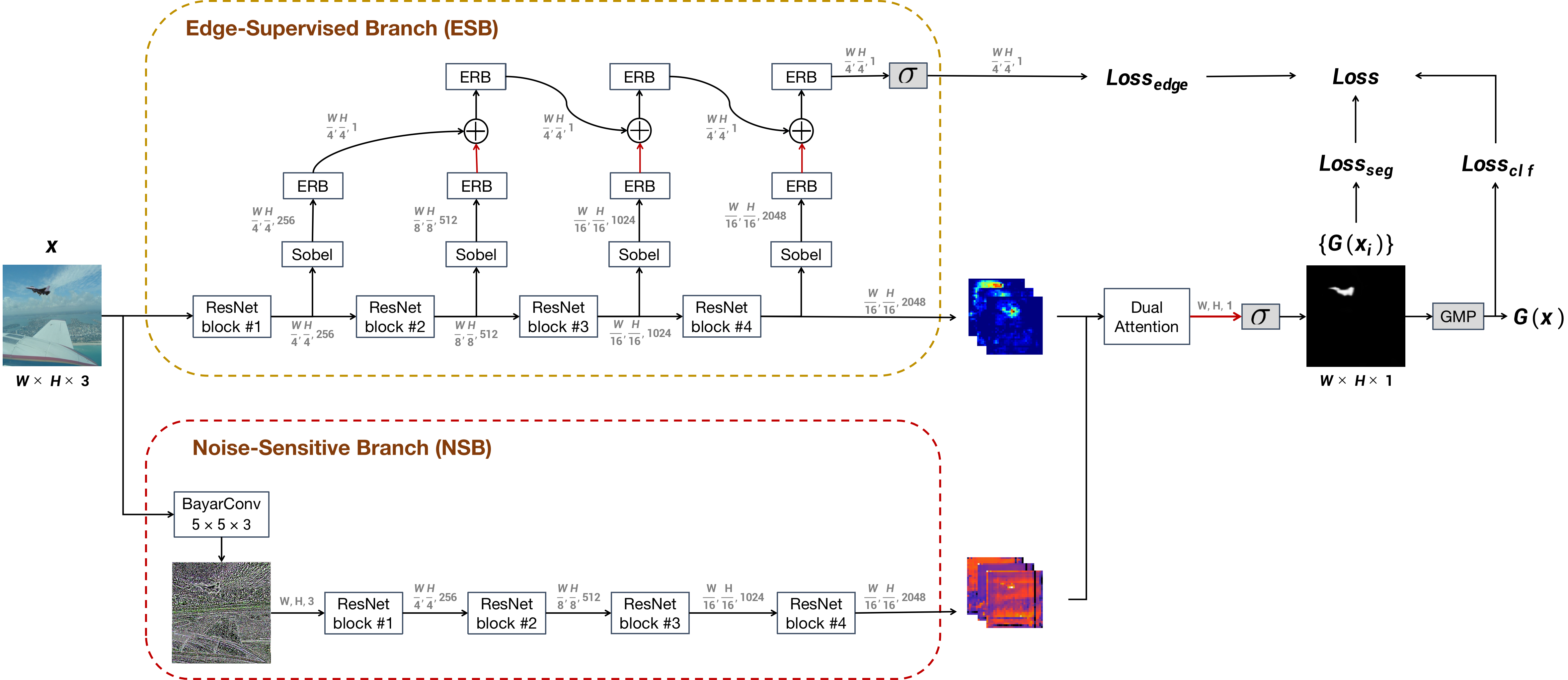}
    \end{center}
    \caption{\textbf{Conceptual diagram of the proposed \model~model}. We use the edge-supervised branch and the noise-sensitive branch to learn semantic-agnostic features for manipulation detection, and multi-scale supervision to strike a balance between model sensitivity and specificity. Non-trainable layers such as sigmoid ($\sigma$) and global max pooling (GMP) are shown in gray.}
    \label{fig:model}
\end{figure*}

Unsurprisingly, the state-of-the-arts are deep learning based \cite{2017MFCN,mantranet,2020GSR,2020Constrained,2020SPAN}, specifically focusing on pixel-level manipulation detection \cite{2017MFCN,mantranet,2020GSR}. With only two classes (\emph{manipulated} versus \emph{authentic}) in consideration, the task appears to be a simplified case of image semantic segmentation.
However, an off-the-shelf semantic segmentation network is suboptimal for the task, as it is designed to capture semantic information, making the network dataset-dependent and do not generalize. Prior research \cite{2020GSR} reports that DeepLabv2 \cite{deeplabv2} trained on the CASIAv2 dataset \cite{casiav2} performs well on the CAISAv1 dataset \cite{casiav1} homologous to CASIAv2, yet performs poorly on the non-homologous COVER dataset \cite{2016COVERAGE}. A similar behavior of FCN \cite{fcn} is also observed in this study. Hence, the key question is how to design and train a deep neural network capable of learning \emph{semantic-agnostic} features that are \emph{sensitive} to manipulations, whilst \emph{specific} to prevent false alarms?

In order to learn semantic-agnostic features, image content has to be suppressed. Depending on at what stage the suppression occurs, we categorize existing methods into two groups, \ie noise-view methods \cite{HPFCN,2020Constrained,2018rgbn,mantranet,2020SPAN} and edge-supervised methods \cite{2017MFCN,2020GSR}. Given the hypothesis that novel elements introduced by slicing and/or inpainting differ from the authentic part in terms of their noise distributions, the first group of methods aim to exploit such discrepancy. The noise map of an input image, generatedfirs either by pre-defined high-pass filters\cite{2012SRM} or by their trainable counterparts~\cite{Bayar-journal,HPFCN}, is fed into a deep network, either alone \cite{HPFCN,2020Constrained} or together with the input image \cite{2018rgbn,mantranet,2020SPAN}. Note that the methods are ineffective for detecting copy-move which introduces no new element. The second group of methods concentrate on finding boundary artifact as manipulation trace around a tampered region, implemented by adding an auxiliary branch to reconstruct the region's edge \cite{2017MFCN,2020GSR}. Note that the prior art \cite{2020GSR} uniformly concatenates features from different layers of the backbone as input of the auxiliary branch. As such, there is a risk that deeper-layer features, which are responsible for manipulation detection, remain semantic-aware and thus not generalizable. 

To measure a model's generalizability, a common evaluation protocol \cite{2017MFCN,2020GSR,mantranet,2020SPAN} is to first train the model on a public dataset, say CASIAv2 \cite{casiav2},  and then test it on other public datasets such as NIST16 \cite{NIST}, Columbia \cite{HsuColumbia}, and CASIAv1 \cite{casiav1}. To our surprise, however, the evaluation is performed exclusively on manipulated images, with metrics w.r.t pixel-level manipulation detection reported. The specificity of the model, which reveals how it handles authentic images and is thus crucial for real-world usability, is ignored. As is shown in Fig. \ref{fig:intro-samples}, their serious false alarm over authentic images leads to unavailability in practical work. In fact, as current methods \cite{mantranet,2020SPAN,2017MFCN} mainly use pixel-wise segmentation losses to which an authentic example can contribute is  marginal, it is nontrivial for these methods to improve their specificity by learning from the authentic.

Inspired by the Border Network \cite{border}, which aggregates features progressively to predict object boundaries, and LesionNet \cite{lesion} that incorporates an image classification loss for retinal lesion segmentation, we propose \textit{multi-view} feature learning with \emph{multi-scale} supervision for image manipulation detection. To the best of our knowledge (Table \ref{tabel:related}), we are the first to jointly exploit the noise view and the boundary artifact to learn manipulation detection features. Moreover, such a joint exploitation is nontrivial. To combine the best of the two worlds, new network structures are needed. Our contributions are as follows: \\
$\bullet$ We propose \model~as a new network for image manipulation detection. As shown in Fig. \ref{fig:model}, \model~contains novel elements designed for learning semantic-agnostic and thus more generalizable features. \\
$\bullet$ We train \model~with multi-scale supervision, allowing us to learn from authentic images, which are ignored by the prior art, and consequently improve the model specificity substantially. \\ 
$\bullet$ Extensive experiments on two training sets and five test sets show that \model~compares favorably against the state-of-the-art. Code and models are available at \url{https://github.com/dong03/MVSS-Net}.

\section{Related Work}\label{sec:relate}



\begin{table*}[htbp]
\begin{center}
\small
\scalebox{0.85}{

\begin{tabular}{l|c|c|c|l|c|c|c}
\toprule
\multicolumn{1}{c|}{\multirow{2}{*}{\textbf{Methods}}} &
  \multicolumn{3}{c|}{\textbf{Views}} &
  \multicolumn{1}{c|}{\multirow{2}{*}{\textbf{Backbone}}} &
  \multicolumn{3}{c}{\textbf{Scales of Supervision}} \\ \cline{2-4} \cline{6-8} 
\multicolumn{1}{c|}{} &
  \textit{RGB} &
  \textit{Noise} &
  \textit{Fusion} &
  \multicolumn{1}{c|}{} &
  \textit{pixel} &
  \textit{edge} &
  \textit{image} \\ \hline
Bappy \etal 2017, J-LSTM\cite{JLSTM}        & + & -                & -                                                                        & Patch-LSTM & + & - & - \\ \hline
Salloum \etal 2017, MFCN\cite{2017MFCN}        & + & -                & -                                                                        & FCN        & + & + & - \\ \hline
Zhou \etal 2020, GSR-Net\cite{2020GSR}      & + & -                & -                                                                        & Deeplabv2  & + & + & - \\ \hline
Li \& Huang 2019, HP-FCN\cite{HPFCN}     & - & High-pass filters & -                                                                        & FCN        & + & - & - \\ \hline
Yang \etal 2020, CR-CNN\cite{2020Constrained}     & - & BayarConv2D      & -                                                                        & Mask R-CNN   & + & - & - \\ \hline
Zhou \etal 2018, RGB-N\cite{2018rgbn}       & + & SRM filter       & \begin{tabular}[c]{@{}c@{}}late fusion\\ (bilinear pooling)\end{tabular} & Faster R-CNN & * & - & - \\ \hline
Wu \etal 2019, ManTra-Net\cite{mantranet} &
  + &\specialcell{SRM filter,\\ BayarConv2D} &
  \specialcell{early fusion\\ (feature concatenation)} &
  Wider VGG &
  + &
  - &
  - \\ \hline
Hu \etal 2020, SPAN\cite{2020SPAN}         & + &\begin{tabular}[c]{@{}c@{}}SRM filter\\ BayarConv2D\end{tabular} 
&  \specialcell{early fusion\\ (feature concatenation)}    & Wider VGG  & + & - & - \\ \hline
\model (\textit{This paper}) & + & BayarConv2D      & \specialcell{late fusion\\ (dual attention)}   & FCN        & + & + & + \\ \bottomrule
\end{tabular}%
}
\end{center}
\caption{\textbf{A taxonomy of the state-of-the-art for image manipulation detection}. Note that edge and image labels used in this work are automatically extracted from pixel-level annotations. So our multi-scale supervision does not use extra manual annotation.}
\label{tabel:related}
\end{table*}

This paper is inspired by a number of recent works that made novel attempts to learn semantic-agnostic features for image manipulation detection, see Table \ref{tabel:related}. In what follows, we describe in brief how these attempts are implemented and explain our novelties accordingly. We focus on deep learning approaches to copy-move / splicing / inpainting detection. For the detection of low-level manipulations such as Gaussian Blur and JPEG compression, we refer to \cite{Bayar-journal}. 

In order to suppress the content information, Li and Huang \cite{HPFCN} propose to implement an FCN's first convolutional layer with trainable high-pass filters and apply their HP-FCN for inpainting detection. Yang \etal  use BayarConv as the initial convolution layer of their CR-CNN \cite{2020Constrained}. Although such constrained convolutional layers are helpful for extracting noise information, using them alone brings in the risk of \del{loosing}\add{losing} other useful information in the original RGB input. Hence, we see an increasing number of works on exploiting information from both the RGB view and the noise view \cite{2018rgbn,mantranet,2020SPAN}. Zhou \etal \cite{2018rgbn} develop a two-stream Faster R-CNN, coined RGB-N, which takes as input the RGB image and its noise counterpart generated by the SRM filter \cite{2012SRM}. Wu \etal \cite{mantranet} and Hu \etal \cite{2020SPAN} use both BayarConv and SRM. Given features from distinct views, the need \del{of}\add{for} feature fusion is on. Feature concatenation at \add{an} early stage is adopted by \cite{mantranet,2020SPAN}. Our \model~is more close to RGB-N as both perform feature fusion at \add{the} late stage. However, different from the non-trainable bilinear pooling used in RGB-N, Dual Attention used in \model~is trainable and is thus more selective.



As manipulating a specific region in a given image inevitably leaves traces between the tampered region and its surrounding, how to exploit such edge artifact also matters for manipulation detection. Salloum \etal develop a multi-task FCN to symmetrically predict a tampered area and its boundary \cite{2017MFCN}. In a more recent work \cite{2020GSR}, Zhou \etal introduce an edge detection and refinement branch which takes as input features at different levels. Given that region segmentation and edge detection are intrinsically two distinct tasks, the challenge lies in how to strike a proper balance between the two. Directly using deeper features for edge detection as done in \cite{2017MFCN} has the risk of affecting the main task of manipulation segmentation, while putting all features together as used in \cite{2020GSR} may let the deeper features be ignored by the edge branch. Our \model~has an edge-supervised branch that effectively resolves these issues.


Last but not least, we observe that the specificity of an image manipulation detector, \ie how it responses to authentic images, is seldom reported. In fact, the mainstream solutions are developed within an image semantic segmentation network. Naturally, they are trained and also evaluated on manipulated images in the context of manipulation segmentation \cite{2020GSR}. The absence of authentic images both in the training and test stages naturally raises concerns regarding the specificity of the detector. In this paper we make a novel attempt to include authentic images for training and test, an important step towards real-world deployment.




\section{Proposed Model}\label{sec:method}

Given an RGB image $x$ of size $W\times H \times 3$, we aim for a multi-head deep network $G$ that not only determines whether the image has been manipulated, but also pinpoints its manipulated pixels. Let $G(x)$ be the network-estimated probability of the image being manipulated. In a similar manner we define $G(x_{i})$ as pixel-wise probability, with $i=1,\ldots,W \times H$. Accordingly, we denote a full-size segmentation map as $\{G(x_{i})\}$. As the image-level decision is naturally subject to pixel-level evidence, we obtain $G(x)$ by Global Max Pooling (GMP) over the segmentation map, \ie 
\begin{equation} \label{eq:gmp}
G(x) \leftarrow \mbox{GMP}\left(\{G(x_{i})\}\right).
\end{equation}

In order to extract generalizable manipulation detection features, we present a new network that accepts both RGB and noise views of the input image. To strike a proper balance between detection sensitivity and specificity, the multi-view feature learning process is jointly supervised by annotations of three scales, \ie pixel, edge and image. 


\subsection{Multi-View Feature Learning} \label{ssec:mvfl}

As shown in Fig. \ref{fig:model},  \model~consists of two branches, with ResNet-50 as their backbones. The edge-supervised branch (ESB) at the top is specifically designed to exploit subtle boundary artifact around tampered regions, whilst the noise-sensitive branch (NSB) at the bottom aims to capture the inconsistency between tampered and authentic regions. Both clues are meant to be semantic-agnostic. 
\subsubsection{Edge-Supervised Branch }\label{sssec:esb}

Ideally, with edge supervision, we hope the response area of the network will be more concentrated on tampered regions. Designing such an edge-supervised network is nontrivial. As noted in Section \ref{sec:relate}, the main challenge is how to construct an appropriate input for the edge detection head. On one hand, directly using features from the last ResNet block is problematic, as this will enforce the deep features to capture low-level edge patterns and consequently affect the main task of manipulation segmentation. While on the other hand, using features from the initial blocks is also questionable, as subtle edge patterns contained in these shallow features can vanish with ease after multiple deep convolutions. A joint use of both shallow and deep features is thus necessary. However, we argue that simple feature concatenation as previously used in \cite{2020GSR} is suboptimal, as the features are mixed and there is no guarantee that the deeper features will receive adequate supervision from the edge head. To conquer the challenge, we propose to construct the input of the edge head in a shallow-to-deep manner.

\begin{figure}[htbp]
\begin{center}
\subfigure[Sobel Layer]{
\begin{minipage}[t]{\linewidth}
\begin{center}
\includegraphics[width=0.8\columnwidth]{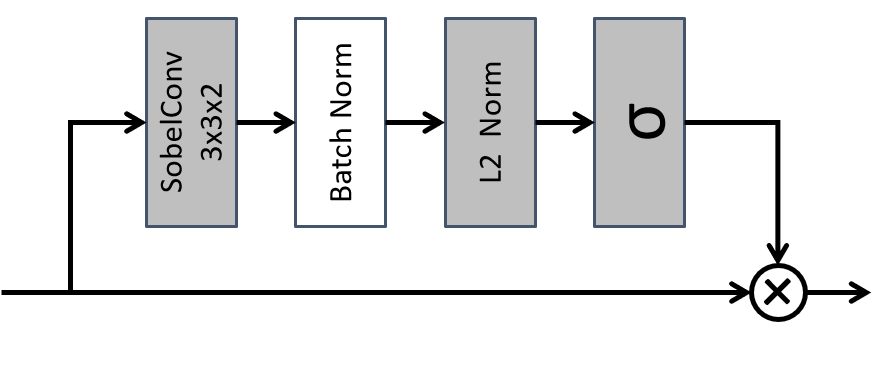}
\end{center}
\label{fig:sobel}
\end{minipage}%
}%

\subfigure[Edge Residual Block (ERB)]{
\begin{minipage}[t]{\linewidth}
\begin{center}
\includegraphics[width=0.8\columnwidth]{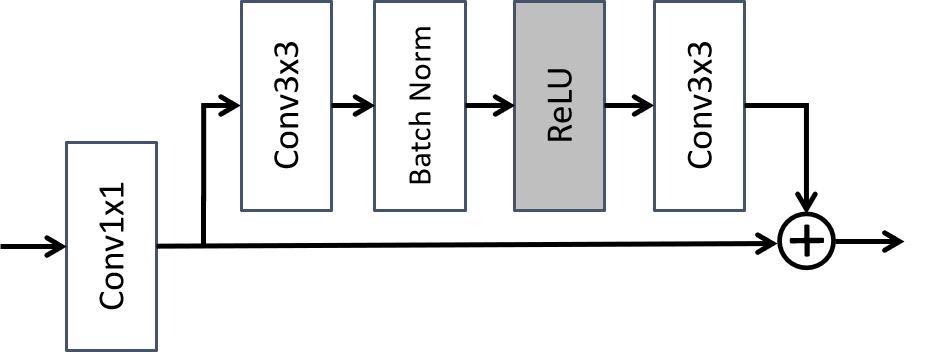}
\end{center}
\label{fig:erb}
\end{minipage}%
}%
\end{center}
\caption{\textbf{Diagrams of (a) Sobel layer and (b) edge residual block}, used in ESB for manipulation edge detection.}
\label{fig:sobel-erb}
\end{figure}
\begin{figure}[htpb]
    \begin{center}
    \includegraphics[width=0.95\columnwidth]{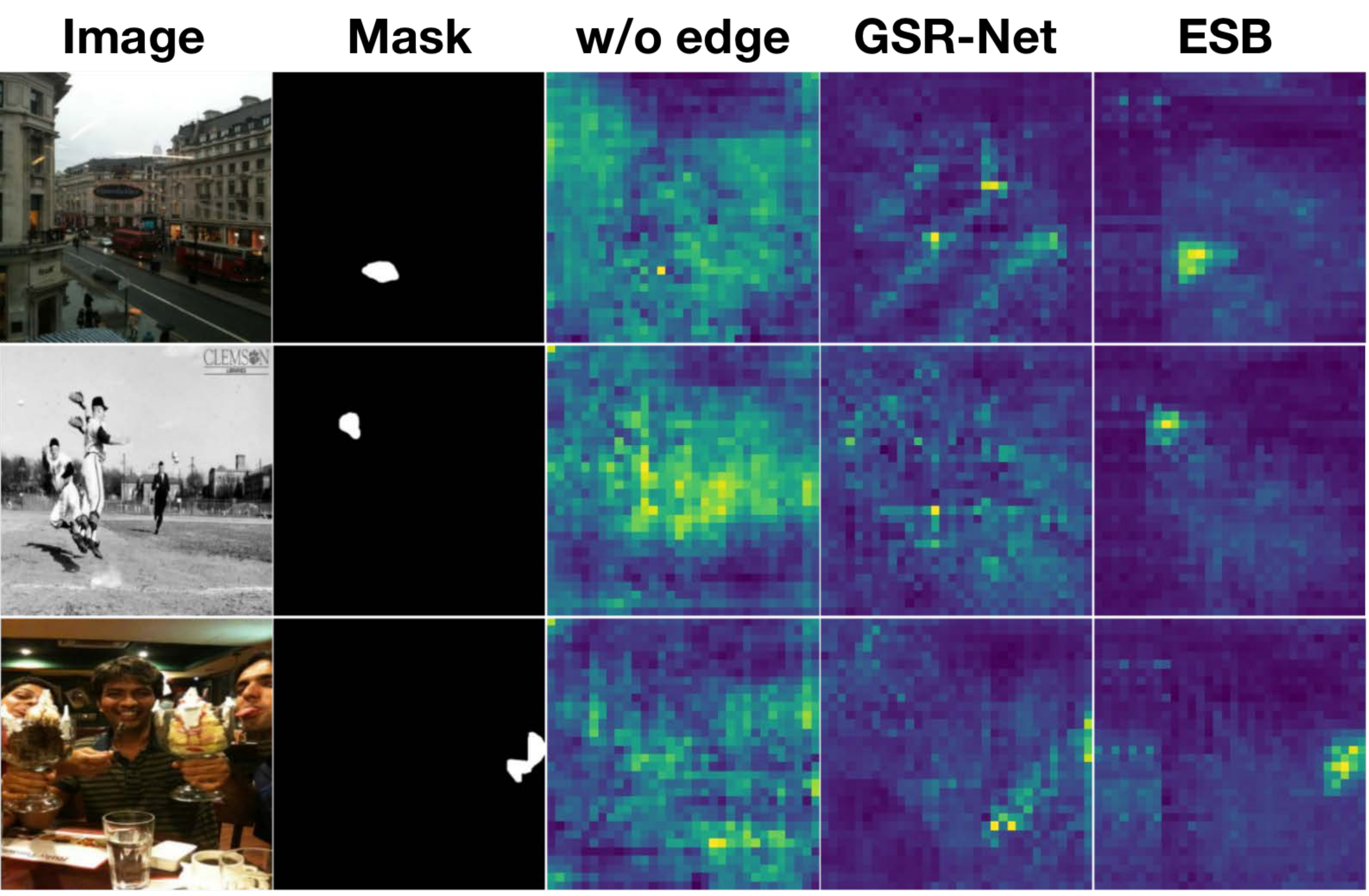}
    \end{center}
    \caption{\textbf{Visualization of averaged feature maps of the last ResNet block}, brighter color indicating \add{a} higher response.  Manipulation from the top to bottom is inpainting, copy-move and splicing. Read from the third column  are \textit{w/o edge}, \ie ResNet without any edge residual block, \textit{GSR-Net}, \ie ResNet with the GSR-Net alike edge branch, and the proposed \textit{ESB}, which produces \add{a} more focused response near tampered regions.}
    \label{fig:visual_esb}
\end{figure}

As illustrated in Fig. \ref{fig:model}, features from different ResNet blocks are combined in a progressive manner for manipulation edge detection. In order to enhance edge-related patterns, we introduce a Sobel layer, see Fig. \ref{fig:sobel}. Features from the $i$-th block first go through the Sobel layer followed by an edge residual block (ERB), see Fig. \ref{fig:erb}, before they are combined (by summation) with their counterparts from the next block. To prevent the effect of accumulation, the combined features go through another ERB (top in Fig. \ref{fig:model}) before the next round of feature combination. We believe such a mechanism helps prevent extreme cases in which deeper features are over-supervised or fully ignored by the edge head. By visualizing feature maps of the last ResNet block in Fig. \ref{fig:visual_esb}, we observe that the proposed ESB indeed produces \add{a} more focused response near tampered regions.  


The output of ESB has two parts: feature maps from the last ResNet block, denoted as $\{f_{esb,1}, \ldots, f_{esb,k}\}$, 
to be used for the main task, and the predicted manipulation edge map, denoted as $\{G_{edge}(x_{i})\}$, obtained by transforming the output of the last ERB with a sigmoid ($\sigma$) layer. The data-flow of this branch is conceptually expressed by Eq. \ref{eq:esb},
\begin{equation}\label{eq:esb}
\left. {\begin{array}{*{20}{r}}
{\left[ {{f_{esb,1}}, \!\ldots ,{f_{esb,k}}} \right]}\\
\{G_{edge}\left( {{x_{i}}} \right)\}
\end{array}} \right\} \!\leftarrow \!\mbox{ERB-ResNet}(x).
\end{equation}


\subsubsection{Noise-Sensitive Branch }\label{sssec:nsb}

In order to fully exploit the noise view, we build a noise-sensitive branch (NSB) parallel to ESB. NSB is implemented as a standard FCN (another ResNet-50 as its backbone). Regarding the choice of noise extraction, we adopt BayarConv \cite{Bayar-journal}, which is found to be better than the SRM filter \cite{2020Constrained}. The output of this branch is an array of $k$ feature maps from the last ResNet block of its backbone, \ie
\begin{equation} \label{eq:nsb}
\{f_{nsb,1}, \ldots, f_{nsb,k}\} \leftarrow \mbox{ResNet}(\mbox{BayarConv}(x)).
\end{equation}

\subsubsection{Branch Fusion by Dual Attention}\label{sssec:fusion}

Given two arrays of feature maps $\{f_{esb,1},\ldots,f_{esb,k}\}$ and $\{f_{nsb,1}, \ldots, f_{nsb,k}\}$ from ESB and NSB, we propose to fuse them by a trainable Dual Attention (DA) module \cite{danet}. This is new, because previous work \cite{2018rgbn} uses bilinear pooling for feature fusion, which is non-trainable.

\begin{figure}[htpb]
    \begin{center}
    \includegraphics[width=\columnwidth]{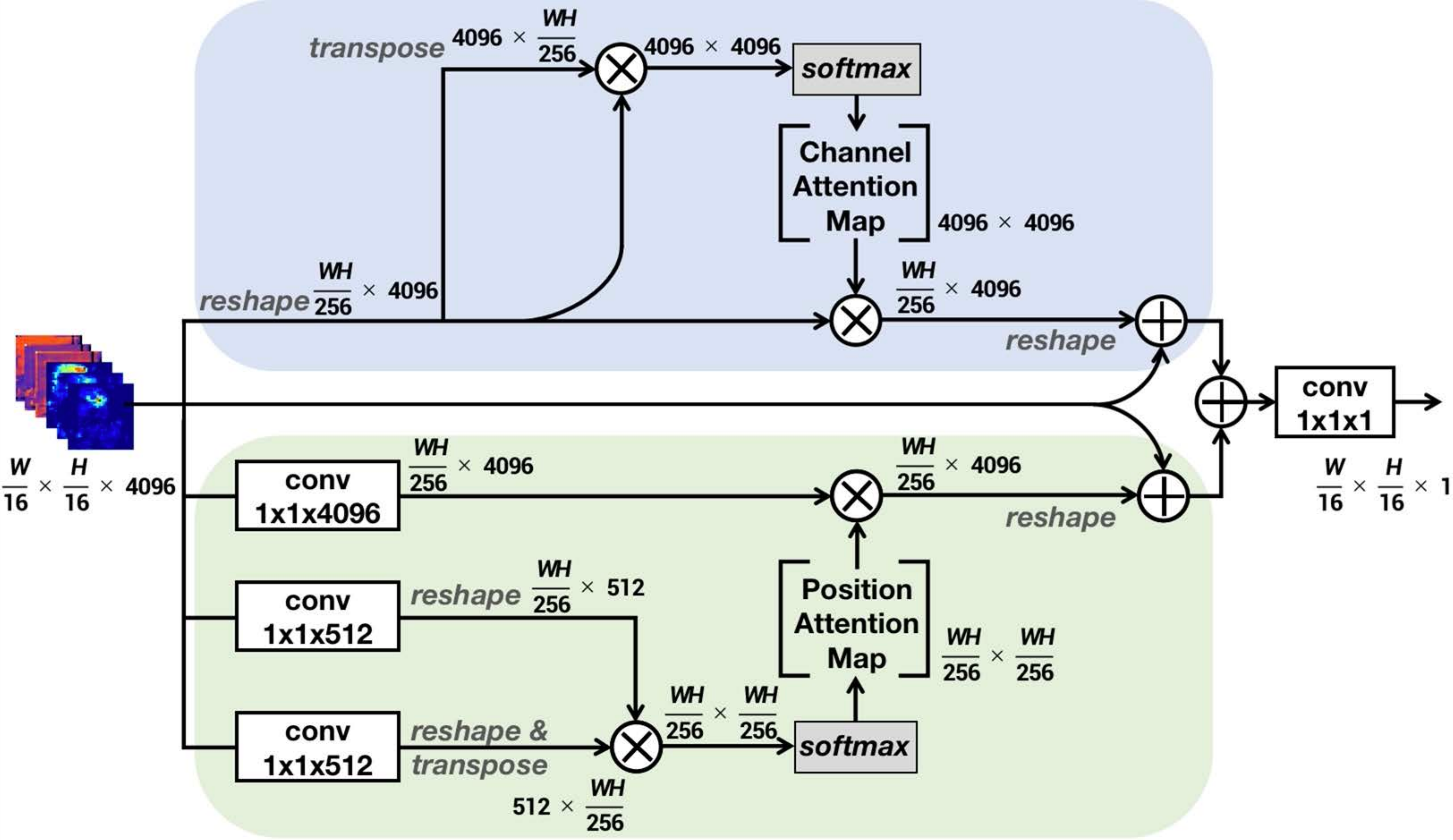}
    \end{center}
    \caption{\textbf{Dual Attention }, with its channel attention module shown in blue and its position attention module shown in green.}
    \label{fig:da}
\end{figure}

The DA module has two attention mechanisms working in parallel: channel attention (CA) and position attention (PA), see Fig. \ref{fig:da}. CA  associates channel-wise  features to selectively emphasize interdependent channel feature maps. Meanwhile, PA selectively updates features at each position by a weighted sum of the features at all positions. The outputs of CA and PA are summed up, and transformed into a feature map of size $\frac{W}{16}\times \frac{H}{16}$, denoted as $\{G^{'}(x_{i})\}$, by a $1 \times 1$ convolution. With parameter-free bilinear upsampling followed by an element-wise sigmoid function, $\{G^{'}(x_{i})\}$ is transformed into the final segmentation map $\{ G(x_{i})\}$. Fusion by dual attention is conceptually expressed as
\begin{equation}
\left\{\! {\begin{array}{*{5}{l}}
\{G^{'}\!(x_{i})\} \!\leftarrow\! DA([f_{esb,1}\!, \!\ldots \!,\!f_{esb,k},\!f_{nsb,1}, \ldots ,f_{nsb,k}]),\\

\{ G(x_{i})\} \!\leftarrow\! \sigma (\mbox{bilinear-upsampling}( \{G^{'}(x_{i})\} )).
\end{array}} \right.\label{eq:da}
\end{equation}


\subsection{Multi-Scale Supervision} \label{ssec:loss} 

We consider losses at three scales, each with its own target, \ie a pixel-scale loss for improving the model's sensitivity for pixel-level manipulation detection, an edge loss for learning semantic-agnostic features and an image-scale loss for improving the model's specificity for image-level manipulation detection. 

\textbf{Pixel-scale loss}. As manipulated pixels are typically in minority in a given image, we use the Dice loss, found to be effective for learning from extremely imbalanced data \cite{lesion}:
\begin{equation} 
loss_{seg}(x) = 1 - \frac{2 \cdot \sum\nolimits_{i = 1}^{W \times H} G(x_i) \cdot y_i}{\sum\nolimits_{i = 1}^{W \times H} G^2(x_i)  + \sum\nolimits_{i = 1}^{W \times H} y^2_i},
\label{eq:dice}
\end{equation}
where $y_i\in\{0,1\}$ is a binary label indicating whether the $i$-th pixel is manipulated.



\textbf{Edge loss}. As pixels of an edge are overwhelmed by non-edge pixels, we again use the Dice loss for manipulation edge detection, denoted as $loss_{edg}$. Since manipulation edge detection is an auxiliary task, we do not compute the $loss_{edg}$ at the full size of $W \times H$. Instead, the loss is computed at a much smaller size of $\frac{W}{4} \times \frac{H}{4}$, see Fig. \ref{fig:model}. This strategy reduces computational cost during training, and in the meanwhile, improves the performance slightly.

\textbf{Image-scale loss}. In order to reduce false alarms, authentic images have to be taken into account in the training stage. This is however nontrivial for the current works \cite{mantranet,HPFCN,2020GSR,2017MFCN} as they all rely on segmentation losses. Consider the widely used binary cross-entropy (BCE) loss for instance. An authentic image with a small percent of its pixels misclassified contributes marginally to the BCE loss, making it difficult to effectively reduce false alarms. Also note that the Dice loss cannot handle the authentic image by definition. Therefore, an image-scale loss is needed. 
%
We adopt the image-scale BCE loss:
\begin{equation} \label{eq:loss-clf}
loss_{clf}(x) = - ( y \cdot \log G(x)  + (1 - y) \cdot \log(1 - G(x)) )
\end{equation}
where $y=\max(\{y_i\})$.

\textbf{Combined loss}. We use a convex combination of the three losses:
\begin{equation}
    Loss = \alpha \cdot loss_{seg} + \beta \cdot loss_{clf} + (1 - \alpha - \beta) \cdot loss_{edg}
\end{equation} 
where $\alpha, \beta \in(0,1)$ are weights. Note that authentic images are only used to compute $loss_{clf}$.

\section{Experiments}\label{sec:eval}
\subsection{Experimental Setup}



\textbf{Datasets}. For the ease of a head-to-head comparison with the state-of-the-art, we adopt CASIAv2 \cite{casiav2} for training and COVER \cite{2016COVERAGE}, Columbia \cite{HsuColumbia}, NIST16 \cite{NIST} and CASIAv1 \cite{casiav1} for testing. Meanwhile, we notice DEFACTO \cite{Mahfoudi2019DEFACTO}, a recently released large-scale dataset, containing 149k images sampled from MS-COCO \cite{Lin2014Microsoft} and auto-manipulated by copy-move, splicing and inpainting. Considering the challenging nature of DEFACTO, we choose to perform our ablation study on this new set. As the set has no authentic images, we construct a training set termed DEFACTO-84k, by randomly sampling 64k positive images from DEFACTO and 20k negative images from MS-COCO. In a similar manner, we build a test set termed DEFACTO-12k, by randomly sampling 6k positive images from the remaining part of DEFACTO and 6k negatives from MS-COCO. Note that to avoid any data leakage, for manipulated images used for training (test), their source images are not included in the test (training) set. In total, our experiments use two training sets and five test sets, see Table \ref{table:dataset}.

\begin{table}[h]
\begin{center}
\scalebox{0.75}{
\begin{tabular}{@{}|l|r|r|r|r|r|@{}}
\hline
\textbf{Dataset} &
  \multicolumn{1}{l|}{\textbf{Negative}} &
  \multicolumn{1}{l|}{\textbf{Positive}} &
  \multicolumn{1}{c|}{\textbf{cpmv}} &
  \multicolumn{1}{c|}{\textbf{spli}} &
  \multicolumn{1}{c|}{\textbf{inpa}} \\ \hline
\textit{Training} & \multicolumn{5}{l|}{}                      \\ \hline
DEFACTO-84k\cite{Mahfoudi2019DEFACTO}       & 20,000 & 64,417 & 12,777 & 34,133 & 17,507 \\ \hline
CASIAv2\cite{casiav2}           & 7,491  & 5,063  & 3,235  & 1,828  & 0      \\ \hline
\textit{Test}     & \multicolumn{5}{l|}{}                      \\ \hline
COVER\cite{2016COVERAGE}             & 100    & 100    & 100    & 0      & 0      \\ \hline
Columbia\cite{HsuColumbia}          & 183    & 180    & 0      & 180    & 0      \\ \hline
NIST16\cite{NIST}            & 0      & 564    & 68     & 288    & 208    \\ \hline
CASIAv1\cite{casiav1}           & 800    & 920    & 459    & 461    & 0      \\ \hline
DEFACTO-12k\cite{Mahfoudi2019DEFACTO}       & 6,000  & 6,000  & 2,000  & 2,000  & 2,000  \\ \hline
\end{tabular}%
}
\end{center}
\caption{\textbf{Two training sets and five test sets in our experiments}. DEFACTO-84k and DEFACTO-12k are used for training and test in the ablation study (Section \ref{ssec:ablation}), while for the SOTA comparison (Section \ref{ssec:eval-sota}) we train on CASIAv2 and evaluate on all test sets.}
\label{table:dataset}
\end{table}



\textbf{Evaluation Criteria}. 
For pixel-level manipulation detection, following previous works \cite{2017MFCN, 2018rgbn, 2020GSR}, we compute pixel-level precision and recall, and report their F1. For image-level manipulation detection, in order to measure the miss detection rate and false alarm rate, we report sensitivity, specificity and their F1. AUC, as a decision-threshold-free metric, is also reported. 
Authentic images per test set 
are only used for image-level evaluation. For both pixel-level and image-level F1 computation, the default threshold is $0.5$, unless otherwise stated.

The overall performance is measured by Com-F1, defined as the harmonic mean of pixel-level and image-level F1. Com-F1 is sensitive to the lowest value of pixel-F1 and image-F1. In
particular, it scores 0 when either pixel-F1 or image-F1 is 0, which does not hold for the arithmetic mean.



\textbf{Implementation}. \model~is implemented in PyTorch and trained on an NVIDIA Tesla V100 GPU. The input size is $512\times512$. The two ResNet-50 used in ESB and NSB are initialized with ImageNet-pretrained counterparts. We use an Adam \cite{2014Adam} optimizer with a learning rate periodically decays from $10^{-4}$ to $10^{-7}$. We set the two weights in the combined loss as $\alpha=0.16$ and $\beta=0.04$, according to the model performance on a held-out validation set from DEFACTO. 
We apply regular data augmentation for training, including flipping, blurring, compression and naive manipulations either by cropping and pasting a squared area or using built-in OpenCV inpainting functions \cite{cv2inp1,cv2inp2}. 



\subsection{Ablation Study}\label{ssec:ablation}
\begin{table*}[htbp]
\begin{center}
\scalebox{0.85}{
\begin{tabular}{l|c|c|c|c|c|c|c|c|c|c|c|r}
\toprule
\multirow{2}{*}{\textbf{Setup}} &
  \multicolumn{3}{c|}{\textbf{Component}} &
  \multicolumn{4}{c|}{\textbf{Pixel-level manipulation detection (F1)}} &
  \multicolumn{4}{c|}{\textbf{Image-level manipulation detection}} &
  \multicolumn{1}{c}{\multirow{2}{*}{\textbf{Com-F1}}} \\ \cline{2-12}
 &
  \textit{loss} &
  \textit{ESB} &
  \textit{NSB} &
  \textit{cpmv.} &
  \textit{spli.} &
  \textit{inpa.} &
  \textit{MEAN} &
  \textit{AUC} &
  \textit{Sen.} &
  \textit{Spe.} &
  \textit{F1} &
  \multicolumn{1}{c}{} \\ \hline
Seg   & - & -         & - & \textbf{0.453} & \textbf{0.722} & \textbf{0.463} & \textbf{0.546} & 0.840 & \textbf{0.827} & 0.620 & 0.709 & 0.617 \\ \hline
Seg+Clf   & + & -         & - & 0.341 & 0.673 & 0.376 & 0.463 & 0.858 & 0.768 & 0.778 & 0.773 & 0.579 \\ \hline
Seg+Clf+N   & + & -         & + & 0.393 & 0.706 & 0.426 & 0.508 & 0.871 & 0.763 & 0.821 & 0.791 & 0.619 \\ \hline
Seg+Clf+E   & + & +         & - & 0.405 & 0.715 & 0.435 & 0.518 & 0.870 & 0.773 & 0.811 & 0.792 & 0.626 \\ \hline
Seg+Clf+E/s   & + & w/o sobel & - & 0.382 & 0.710 & 0.422 & 0.505 & 0.869 & 0.792 & 0.789 & 0.790 & 0.616 \\ \hline
Seg+Clf+G   & + & GSR-Net    & - & 0.363 & 0.714 & 0.421 & 0.499 & 0.864 & 0.813 & 0.779 & 0.796 & 0.613 \\ \hline
Full setup   & + & + & + & 0.446 & 0.714 & 0.455 & 0.538 & \textbf{0.886} & 0.797 & 0.802 & \textbf{0.799} & \textbf{0.643} \\\hline
Ensemble(N, E) & + & + & + & 0.384 & 0.708 & 0.437 & 0.510 & 0.878 & 0.731 & \textbf{0.876} & 0.797 & 0.622 \\ 
\bottomrule
\end{tabular}%
}

\end{center}
\caption{\textbf{Ablation study of \model}. Training: DEFACTO-84k. Test: DEFACTO-12k. Copy-move, splicing and inpainting are shorten\add{ed} as \textit{cmpv}, \textit{spli} and \textit{inpa}, respectively. Best number per column is shown in \textbf{bold}. The top performance of the full setup justifies the necessity of the individual components used in \model.}
\label{table:ablation}
\end{table*}

For revealing the influence of the individual components, we evaluate the performance of the proposed model in varied setups with the components added progressively.


We depart from FCN-16 without multi-view multi-scale supervision. Recall that we use a DA module for branch fusion. So for a fair comparison, we adopt FCN-16 with DA, making it essentially an implementation of DANet~\cite{danet}. The improved FCN-16 scores better than its standard counterpart, \eg UNet~\cite{unet}, DeepLabv3~\cite{deeplabv3} and DeepLabv3+~\cite{deeplab}, see the supplement. 
This competitive baseline is referred to as \emph{Seg} in Table \ref{table:ablation}.  

\begin{figure}[htpb]
    \begin{center}
    \includegraphics[width=0.95\columnwidth]{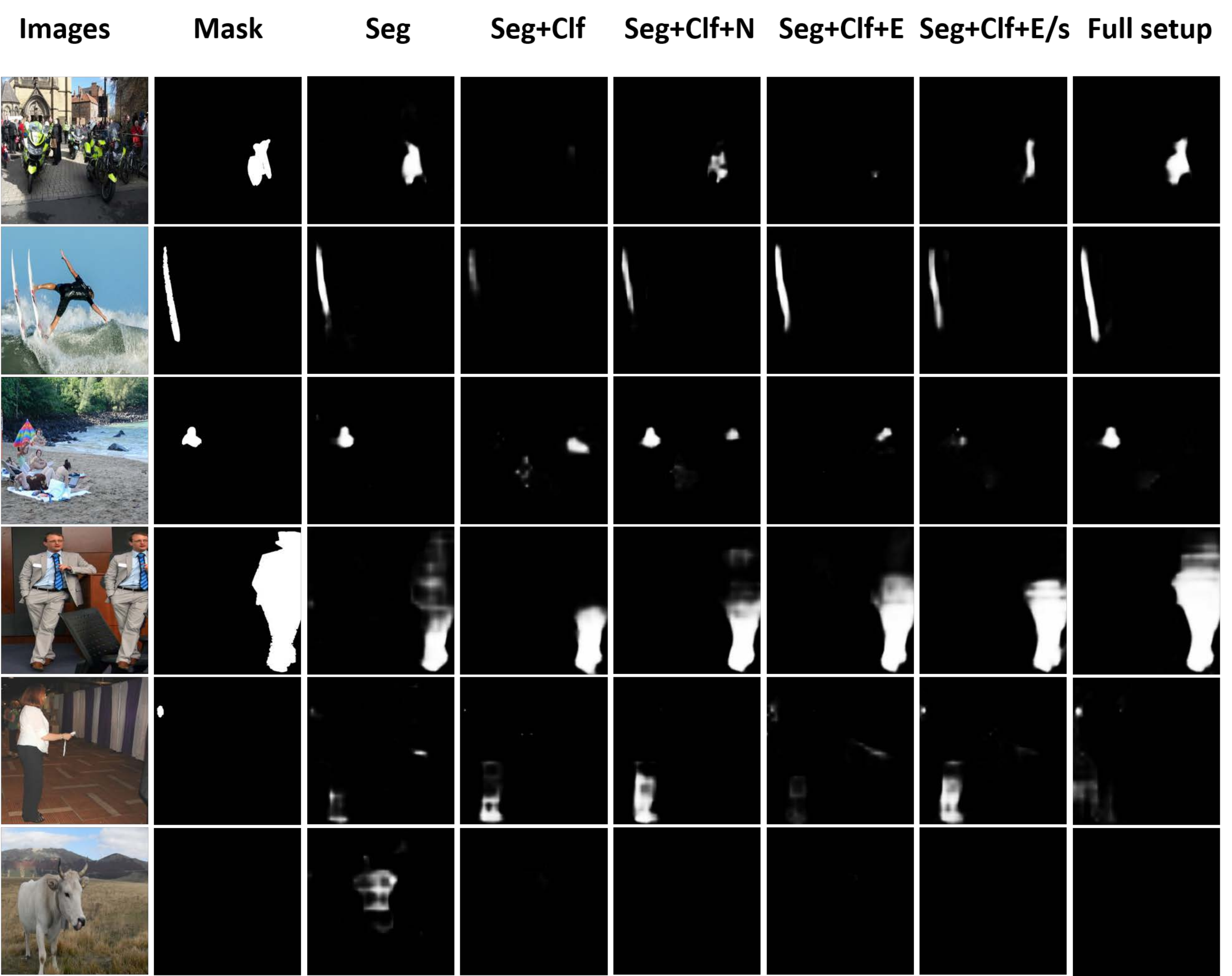}
    \end{center}
    \caption{\textbf{Pixel-level manipulation detection results of \model~in varied setups}. The test image in the last row is authentic. }
    \label{fig:ab_example}
\end{figure}

\textbf{Influence of the image classification loss}. Comparing \emph{Seg+Clf} and \emph{Seg}, we see a clear increase in specificity and a clear drop in sensitivity, suggesting that adding $loss_{clf}$ makes the model more conservative for reporting manipulation. This change is not only confirmed by lower pixel-level performance, but is also  observed in the fourth column of Fig. \ref{fig:ab_example}, showing that manipulated areas predicted by \emph{Seg+Clf} are much reduced.

\textbf{Influence of NSB}. Since \del{Setup\#2}\add{\emph{Seg+Clf+N}} is obtained by adding NSB into \del{Setup\#1}\add{\emph{Seg+Clf}}, its better performance verifies the effectiveness of NSB for improving manipulation detection at both pixel-level and image-level. 


\textbf{Influence of ESB}. The better performance of \del{Setup\#3}\add{\emph{Seg+Clf+E}} against \del{Setup\#1}\add{\emph{Seg+Clf}} justifies the effectiveness of ESB. \del{Setup\#4}\add{\emph{Seg+Clf+E/s}} is obtained by removing the Sobel operation from \del{Setup\#3}\add{\emph{Seg+Clf+E}}, so its performance degeneration in particular on copy-move detection (from 0.405 to 0.382, \textit{cmpv} in Table \ref{table:ablation}) indicates the necessity of this operation. 


\textbf{ESB versus GSR-Net}. \del{Setup\#5}\add{\emph{Seg+Clf+G}} is obtained by replacing our ESB with the edge branch of GSR-Net. The overall performance of \del{Setup\#5}\add{\emph{Seg+Clf+G}} is lower than \del{Setup\#3}\add{\emph{Seg+Clf+E}}. Moreover, there is a larger performance gap on \emph{cmpv} (ESB of 0.405 versus GSR-Net of 0.363). The results clearly demonstrate the superiority of the proposed ESB over the prior art.


\textbf{Influence of two branch fusion}. The full setup, with ESB and NSB fused by dual attention, performs the best, showing the complementarity of the individual components. To further justify the necessity of our dual attention based fusion, we make an alternative solution which ensembles \emph{Seg+Clf+N} and \emph{Seg+Clf+E} by model averaging, refereed to as \emph{Ensemble(N,E)}. 
The full setup is better than \emph{Ensemble(N,E)}, showing the advantage of our fusion method\footnote{Comparison to fusion by bilinear pooling is in the supplement.}. 


Fig. \ref{fig:ab_example} shows 
some qualitative results. From the left to right, the results demonstrate how \model~strikes a good balance between sensitivity and specificity. 
Note that the best pixel-level performance of FCN is due to the fact that the training and test sets are homologous. Next, we evaluate the generalizability of FCN and \model. 

\begin{table*}[htbp]
\renewcommand{\arraystretch}{1.05}
\begin{center}

\small
\setlength{\tabcolsep}{0.7mm}

\scalebox{0.85}{
\begin{tabular}{l|cccccc|cccccc}
\toprule
\multirow{2}{*}{\textbf{Method}} & \multicolumn{6}{c|}{\textbf{Optimal threshold per model \& testset}} & \multicolumn{6}{c}{\textbf{Fixed threshold (0.5)}}  \\
\cline{2-13}
& \textit{NIST} & \textit{Columbia} & \textit{CASIAv1}  & \textit{COVER}  & \textit{DEFACTO-12k}  & \textit{MEAN} & \textit{NIST} & \textit{Columbia} & \textit{CASIAv1} & \textit{COVER}    & \textit{DEFACTO-12k}    & \textit{MEAN}  \\ \hline
MFCN\cite{2017MFCN}                   
& 0.422    & 0.612       & 0.541      & n.a.     & n.a.      & n.a.    & n.a.  & n.a.     & n.a.    & n.a.  & n.a.    & n.a   \\
RGB-N\cite{2018rgbn}                  
& n.a.     & n.a.        & 0.408      & 0.379    & n.a.      & n.a.    & n.a.  & n.a.     & n.a.    & n.a.  & n.a.    & n.a   \\
HP-FCN\cite{HPFCN}               
& 0.360    & 0.471       & 0.214      & 0.199    & 0.136     & 0.276   & 0.121 & 0.067    & 0.154   & 0.003 & 0.055   & 0.080 \\
ManTra-Net\cite{mantranet}              
& 0.455    & \textbf{0.709}       & 0.692      & 0.772    & \textbf{0.618}     & 0.649   & 0.000 & 0.364    & 0.155   & 0.286 & \textbf{0.155}   & 0.192 \\
CR-CNN\cite{2020Constrained}
& 0.428    & 0.704       & 0.662      & 0.470    & 0.340     & 0.521   & 0.238 & 0.436    & 0.405   & 0.291 & 0.132   & 0.300 \\
GSR-Net\cite{2020GSR}                 
& 0.456    & 0.622       & 0.574      & 0.489    & 0.379     & 0.504   & 0.283 & 0.613    & 0.387   & 0.285 & 0.051   & 0.324 \\\midrule
FCN                 
& 0.507    & 0.586       & 0.742      & 0.573    & 0.401     & 0.562   & 0.167 & 0.223    & 0.441   & 0.199 & 0.130   & 0.232 \\
\model                  
& \textbf{0.737}    & 0.703       & \textbf{0.753}      &  \textbf{0.824}    & 0.572     &  \textbf{0.718}   & \textbf{0.292} & \textbf{0.638}    & \textbf{0.452}   & \textbf{0.453} & 0.137   & \textbf{0.394} \\ \bottomrule
\end{tabular}
}
\end{center}
\caption{\textbf{Performance of pixel-level manipulation detection}. Best result per test set is shown in bold. All the models are trained on CASIAv2, except for ManTra-Net and HP-FCN.}
\label{table:pix-f1}
\end{table*}


\subsection{Comparison with State-of-the-art} \label{ssec:eval-sota}


\begin{table*}[htpb]
\renewcommand{\arraystretch}{1.05}
\begin{center}
\small
\setlength{\tabcolsep}{1.2mm}
\scalebox{0.85}{
\begin{tabular}{l|cccc|cccc|cccc|cccc}
\toprule
\multicolumn{1}{c|}{\multirow{2}{*}{\textbf{Method}}} &
  \multicolumn{4}{c|}{\textbf{Columbia}} &
  \multicolumn{4}{c|}{\textbf{CASIAv1}} &
  \multicolumn{4}{c|}{\textbf{COVER}} &
  \multicolumn{4}{c}{\textbf{DEFACTO-12k}} \\ \cline{2-17} 
\multicolumn{1}{c|}{} &
  \textit{AUC} &
  \textit{Sen.} &
  \textit{Spe.} &
  \textit{F1} &
  \textit{AUC} &
  \textit{Sen.} &
  \textit{Spe.} &
  \textit{F1} &
  \textit{AUC} &
  \textit{Sen.} &
  \textit{Spe.} &
  \textit{F1} &
  \textit{AUC} &
  \textit{Sen.} &
  \textit{Spe.} &
  \textit{F1} \\ \hline
ManrTra-Net\cite{mantranet} & 0.701 & 1.000 & 0.000 & 0.000 & 0.141 & 1.000 & 0.000 & 0.000 & 0.491 & 1.000 & 0.000 & 0.000 & 0.543 & 1.000 & 0.000 & 0.000 \\
CR-CNN\cite{2020Constrained}     & 0.783 & 0.961 & 0.246 & 0.392 & 0.766 & 0.930 & 0.224 & 0.361 & 0.566 & 0.967 & 0.070 & 0.131 & 0.567 & 0.774 & 0.267 & 0.397 \\
GSR-Net\cite{2020GSR}     & 0.502 & 1.000 & 0.011 & 0.022 & 0.502 & 0.994 & 0.011 & 0.022 & 0.515 & 1.000 & 0.000 & 0.000 & 0.456 & 0.914 & 0.001 & 0.002 \\\midrule
FCN     & 0.762 & 0.950 & 0.322 & 0.481 & 0.796 & 0.717 & 0.844 & \textbf{0.775} & 0.541 & 0.900 & 0.100 & 0.180 & 0.551 & 0.711 & 0.338 & \textbf{0.458} \\
\model   & \textbf{0.980} & 0.669 & 1.000 & \textbf{0.802} & \textbf{0.839} & 0.615 & 0.969 & 0.752 & \textbf{0.731} & 0.940 & 0.140  & \textbf{0.244} & \textbf{0.573} & 0.817 & 0.268 & 0.404 \\ \bottomrule
\end{tabular}%
}
\end{center}
\caption{\textbf{Performance of image-level manipulation detection on Columbia, CASIAv1, COVER and DEFACTO-12k}. \textit{Sen.}:  sensitivity. \textit{Spe.}:  specificity. NIST16, which has no authentic images, is excluded. The default decision threshold of $0.5$ is used for all models.}
\label{table:perf-img}
\end{table*}

\begin{table}[htpb]
\begin{center}
\renewcommand{\arraystretch}{1}
\small
\setlength{\tabcolsep}{1.2mm}
\scalebox{0.85}{
\begin{tabular}{l|c|c|c|c}
\hline
\textbf{Method} &
  \textbf{Columbia} &
  \textbf{CASIAv1} &
  \textbf{COVER} &
  \textbf{DEFACTO-12k} \\ \hline
ManrTra-Net\cite{mantranet} & 0.000 & 0.000 & 0.000 & 0.000 \\
CR-CNN\cite{2020Constrained}     & 0.413 & 0.382 & 0.181 & 0.198 \\
GSR-Net\cite{2020GSR}     & 0.042 & 0.042 & 0.000 & 0.004 \\\hline
FCN     & 0.305 & 0.562 & 0.189 & 0.203 \\
\model   & \textbf{0.711} & \textbf{0.565} & \textbf{0.317} & \textbf{0.205} \\ \hline
\end{tabular}%
}
\end{center}
\caption{\textbf{Com-F1, the harmonic mean of pixel-level F1 and image-level F1, on four test sets}. }
\label{table:perf-img-comf1}
\end{table} 

\textbf{Baselines}. For a fair and reproducible comparison, we have to be selective, choosing the state-of-the-art that meets one of the following three criteria: 1) pre-trained models  released by paper authors, 2) source code  publicly available, or 3) following a common evaluation protocol where CASIAv2 is used for training and other public datasets are used for test\add{ing}. Accordingly, we compile a list of six published baselines as follows:\\
$\bullet$ Models available: HP-FCN \cite{HPFCN}, trained on a private set of inpainted images\footnote{\url{https://github.com/lihaod/Deep_inpainting_localization}}, ManTra-Net~\cite{mantranet}, trained on a private set of millions of manipulated images\footnote{\url{https://github.com/ISICV/ManTraNet}}, and CR-CNN~\cite{2020Constrained}, trained on CASIAv2\footnote{\url{https://github.com/HuizhouLi/Constrained-R-CNN}}. We use these models directly. \\
$\bullet$ Code available: GSR-Net \cite{2020GSR}, which we train using author-provided  code\footnote{\url{https://github.com/pengzhou1108/GSRNet}}. We cite their results where appropriate and use our re-trained model only when necessary. \\
$\bullet$ Same evaluation protocol: MFCN \cite{2017MFCN}, RGB-N \cite{2018rgbn} with numbers quoted from the same team \cite{2020GSR}. 

We re-train FCN (\emph{Seg}) and \model (\emph{full setup}) from scratch on CASIAv2.

\begin{figure}[htbp]
\begin{center}

\subfigure[Performance curves w.r.t. JPEG compression]{
\begin{minipage}[t]{\linewidth}
\includegraphics[width=8.3cm]{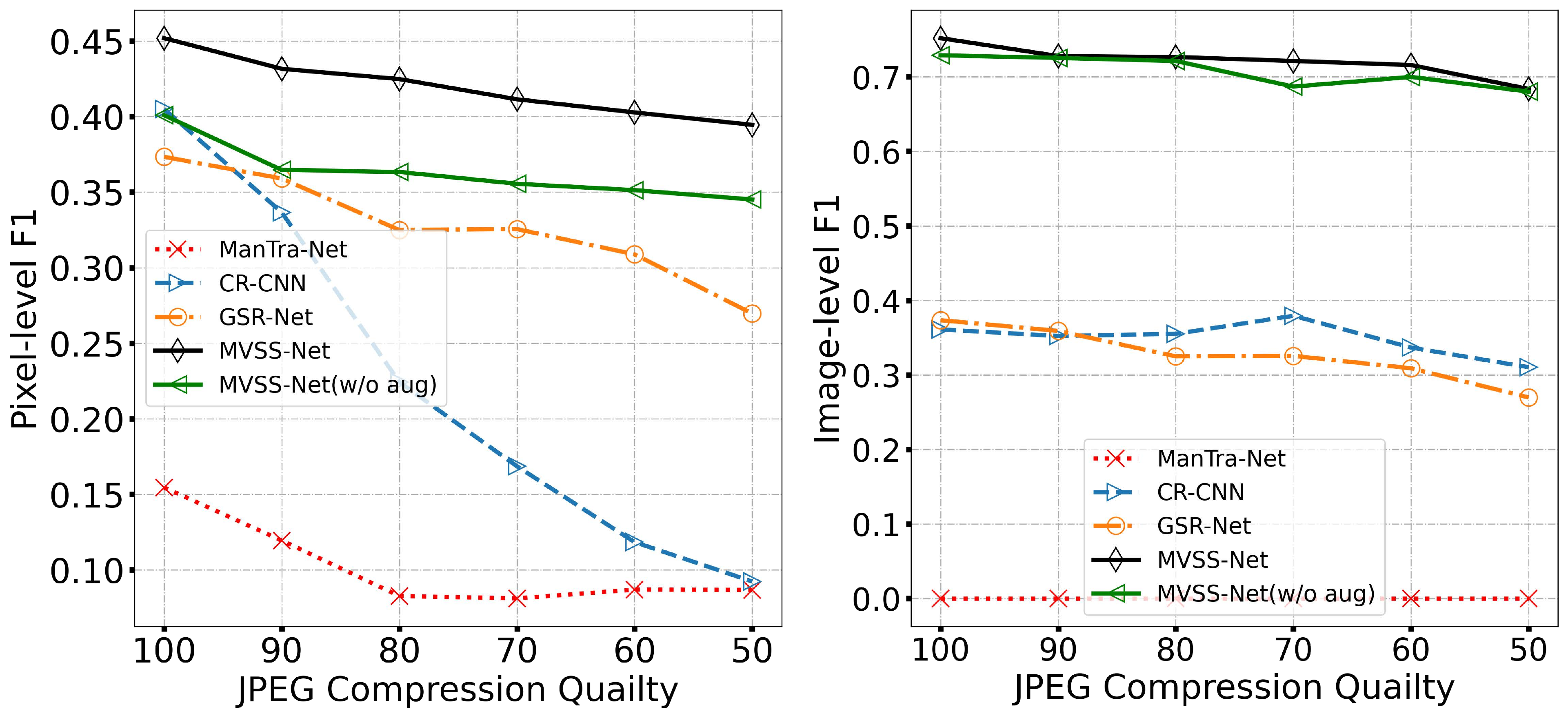}
\end{minipage}%
}%

\subfigure[Performance curves w.t.r. Gaussian Blurs]{
\begin{minipage}[t]{\linewidth}
\includegraphics[width=8.3cm]{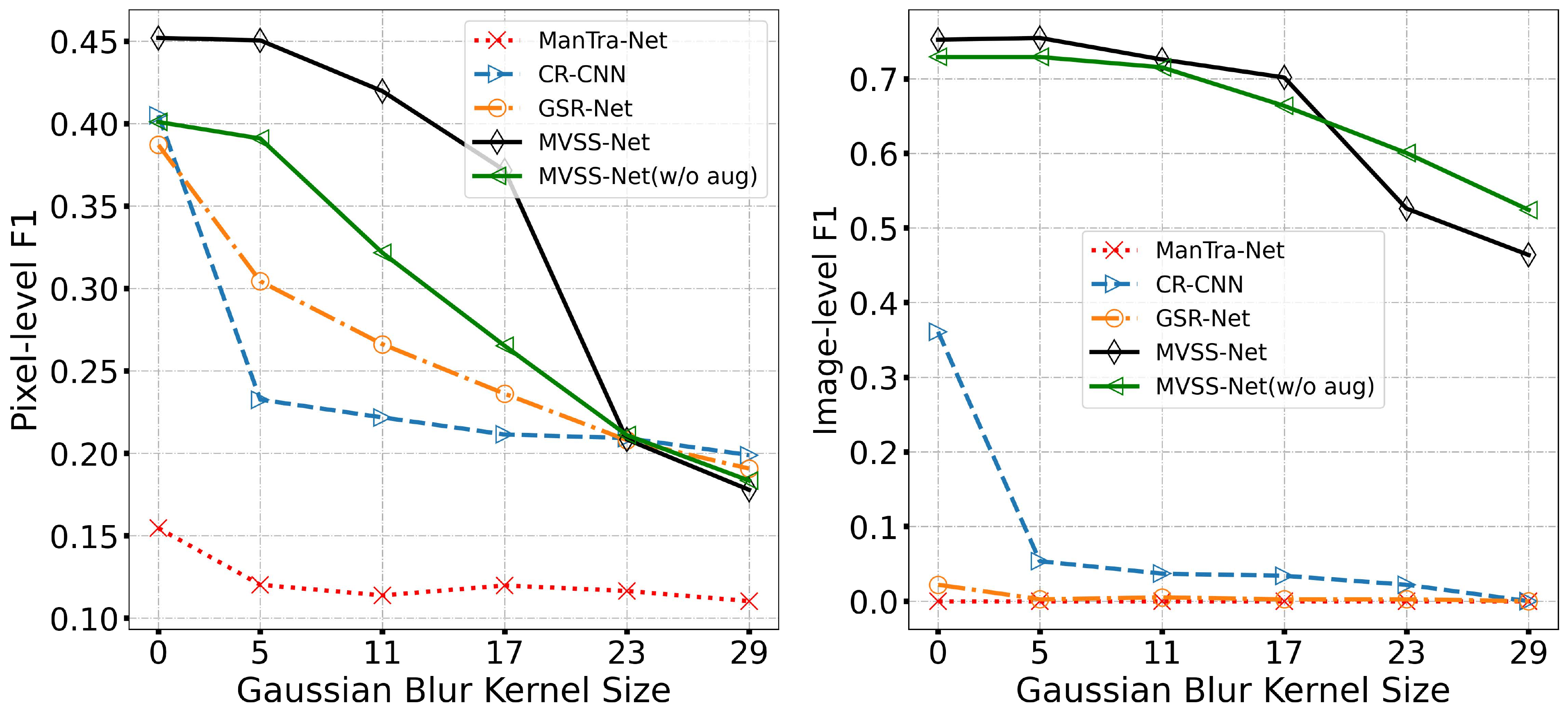}
\end{minipage}
}%
\end{center}
\caption{\textbf{Robustness evaluation against JPEG compression and Gaussian Blurs on CASIAv1}.}

\label{fig:defacto_robust}
\end{figure}

\textbf{Pixel-level manipulation detection}. 
The performance of distinct models is given in Table \ref{table:pix-f1}. \model~is the best in terms of overall performance. We attribute the clearly better performance of ManTra-Net on DEFACTO-12k to its large-scale training data, which was also originated from MS-COCO as DEFACTO-12k. As \model~is derived from FCN, its superior performance in this cross-dataset setting justifies its better generalizability.

As HP-FCN is specially designed for inpainting detection, we narrow down the comparison to detecting  the inpainting subsets in NIST16 and DEFACTO-12k. Again, \model~outperforms HP-FCN: 0.565 versus 0.284 on NIST16 and 0.391 versus 0.106 on DEFACTO-12k.

\textbf{Image-level manipulation detection}.
 Table \ref{table:perf-img} shows the performance of distinct models, all using the default decision threshold of 0.5. \model~is again the top performer. With its capability of learning from authentic images, \model~obtains higher specificity (and thus lower false alarm rate) on most test sets. Our model also has the best AUC scores, meaning it is better than the baselines on a wide range of operation points. 
 


The overall performance on both pixel-level and image-level manipulation detection is provided in Table \ref{table:perf-img-comf1}.



\textbf{Robustness evaluation}. 
JPEG compression and Gaussian blur are separately applied on CASIAv1. 
ManTra-Net used a wide range of data augmentations including compression, while CR-CNN and GSR-Net did not
use such data augmentation. So for a more fair comparison, we also train \model~ with compression and blurring excluded from data augmentation, denoted as \model~(w/o aug). Performance curves in Fig. \ref{fig:defacto_robust} show better robustness of \model~ and \model~(w/o aug).






\textbf{Efficiency test}. 
%
We measure the inference efficiency in terms of frames per second (FPS) . Tested on NVIDIA Tesla V100 GPU, CR-CNN, ManTra-Net and GSR-Net run at FPS of 3.1, 2.8 and 31.7, respectively. 
\model~runs at FPS of 20.1,  sufficient for real-time application. 



\section{Conclusions}
Our image manipulation detection experiments on five benchmark sets allow us to draw the following conclusions. For learning semantic-agnostic features, both noise and edge information are helpful, whilst the latter is better when used alone. For exploiting the edge information, our proposed edge-supervised branch (ESB) is more effective than the previously used feature concatenation. ESB steers the network to be more concentrated on tampered regions. Regarding the specificity of manipulation detection, we empirically show that the state-of-the-arts suffer from poor specificity. The inclusion of the image classification loss improves the specificity, yet at the cost of a clear performance drop for pixel-level manipulation detection. Multi-view feature learning has to be used together with multi-scale supervision. The resultant \model~is a new state-of-the-art for image manipulation detection.


\medskip
\textbf{Acknowledgements}. This research was supported by NSFC (U1703261), BJNSF  (4202033), the Fundamental Research Funds for the Central Universities and the Research Funds of Renmin University of China (No. 18XNLG19), and Public Computing Cloud, Renmin University of China. This work was initially inspired by the Security AI Challenge: Forgery Detection on Certificate Image, Alibaba Security.

{\small

\bibliographystyle{ieee_fullname}
}

\end{document}